\documentclass[sigconf]{acmart}
\usepackage{orcidlink}
\usepackage{algorithm}
\usepackage{algorithmic}
\usepackage{multirow}
\usepackage{float}

\newcommand{\eg}{e.g.}

\copyrightyear{2025}
\acmYear{2025}
\setcopyright{acmlicensed}\acmConference[MM '25]{Proceedings of the 33rd ACM International Conference on Multimedia}{October 27--31, 2025}{Dublin, Ireland}
\acmBooktitle{Proceedings of the 33rd ACM International Conference on Multimedia (MM '25), October 27--31, 2025, Dublin, Ireland}
\acmDOI{10.1145/3746027.3755263}
\acmISBN{979-8-4007-2035-2/2025/10}
\begin{document}

%%
%% The "title" command has an optional parameter,
%% allowing the author to define a "short title" to be used in page headers.
\title{DeflareMamba: Hierarchical Vision Mamba for Contextually Consistent Lens Flare Removal}

%%
%% The "author" command and its associated commands are used to define
%% the authors and their affiliations.
%% Of note is the shared affiliation of the first two authors, and the
%% "authornote" and "authornotemark" commands
%% used to denote shared contribution to the research.
\author{Yihang Huang}
\orcid{0009-0003-2986-7484}
\affiliation{%
  \institution{School of Artificial Intelligence, Beijing Normal University}
  \city{Beijing}
  \country{China}
}

\author{Yuanfei Huang}
\orcid{0000-0002-5242-9904}
\authornote{Corresponding author. Email: yfhuang@bnu.edu.cn}
\affiliation{%
	\institution{School of Artificial Intelligence, Beijing Normal University}
	\institution{Engineering Research Center of Intelligent Technology and Educational Application, Ministry of Education}
	\city{Beijing}
	\country{China}
}

\author{Junhui Lin}
\orcid{0009-0006-8661-2869}
\affiliation{%
	\institution{School of Artificial Intelligence, Beijing Normal University}
	\city{Beijing}
	\country{China}
}

\author{Hua Huang}
\orcid{0000-0003-2587-1702}
\affiliation{%
	\institution{School of Artificial Intelligence, Beijing Normal University}
	\institution{Engineering Research Center of Intelligent Technology and Educational Application, Ministry of Education}
	\city{Beijing}
	\country{China}
}

%%
%% By default, the full list of authors will be used in the page
%% headers. Often, this list is too long, and will overlap
%% other information printed in the page headers. This command allows
%% the author to define a more concise list
%% of authors' names for this purpose.
\renewcommand{\shortauthors}{Yihang Huang, Yuanfei Huang, Junhui Lin, Hua Huang.}

%%
%% The abstract is a short summary of the work to be presented in the
%% article.
\begin{abstract}
  Lens flare removal remains an information confusion challenge in the underlying image background and the optical flares, due to the complex optical interactions between light sources and camera lens. While recent solutions have shown promise in decoupling the flare corruption from image, they often fail to maintain contextual consistency, leading to incomplete and inconsistent flare removal. To eliminate this limitation, we propose DeflareMamba, which leverages the efficient sequence modeling capabilities of state space models while maintains the ability to capture local-global dependencies. Particularly, we design a hierarchical framework that establishes long-range pixel correlations through varied stride sampling patterns, and utilize local-enhanced state space models that simultaneously preserves local details. To the best of our knowledge, this is the first work that introduces state space models to the flare removal task. Extensive experiments demonstrate that our method effectively removes various types of flare artifacts, including scattering and reflective flares, while maintaining the natural appearance of non-flare regions. Further downstream applications demonstrate the capacity of our method to improve visual object recognition and cross-modal semantic understanding. Code is available at \href{https://github.com/BNU-ERC-ITEA/DeflareMamba}{https://github.com/BNU-ERC-ITEA/DeflareMamba}.
\end{abstract}

%%
%% The code below is generated by the tool at http://dl.acm.org/ccs.cfm.
%% Please copy and paste the code instead of the example below.
%%

%\ccsdesc[300]{Computing methodologies~Computational photography}

%%
%% Keywords. The author(s) should pick words that accurately describe
%% the work being presented. Separate the keywords with commas.
%% A "teaser" image appears between the author and affiliation
%% information and the body of the document, and typically spans the
%% page.
\begin{teaserfigure}
	\centering
  \includegraphics[width=\textwidth, height=0.195\linewidth]{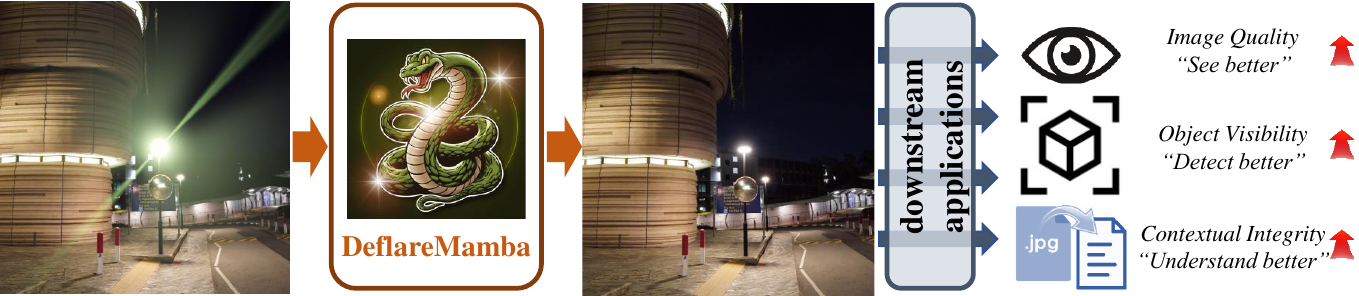}
  \caption{DeflareMamba effectively enhances the quality of image media, thereby enabling downstream applications that benefit from improved image quality, increased object visibility, and enhanced contextual integrity.}
  \label{fig:teaser}
\end{teaserfigure}

%%
%% This command processes the author and affiliation and title
%% information and builds the first part of the formatted document.
\maketitle

\section{Introduction}
%%主要写了flare的定义和它对多媒体用户体验的影响，多模态对其的影响，以此引出flare removal的重要性  
Lens flare~\cite{hullin2011physically,li2021let,reddy2021lens} fundamentally represents a parasitic signal that nonlinearly couples with the intrinsic image content. This coupling leads to significant degradation in the quality of images and videos. 
% Firstly, it severely impacts the visual quality and user experience by introducing undesired artifacts, \eg, luminance patterns and color shifts. Besides, in cross-modal alignment tasks such as image-text or video-audio pairs, these artifacts can introduce semantic noise that disrupts the alignment between different modalities.
Firstly, it severely compromises the visual quality and deteriorates user experience by introducing undesired artifacts, \eg, irregular luminance patterns and color shifts. Besides, these artifacts can also degrade the performance of downstream vision tasks like object detection and recognition. Moreover, in cross-modal tasks, such as image-text or video-audio alignment, these artifacts may introduce semantic noise that disrupts the correspondence between modalities. 
In Figure~\ref{fig:teaser}, it is crucial to decoupling these optical artifacts from the intrinsic scene information, thereby ensuring clean input data for multimedia and multimodal applications~\cite{nussberger2015robust}.

\begin{figure}[!t]
	\centering
	\includegraphics[width=1\linewidth]{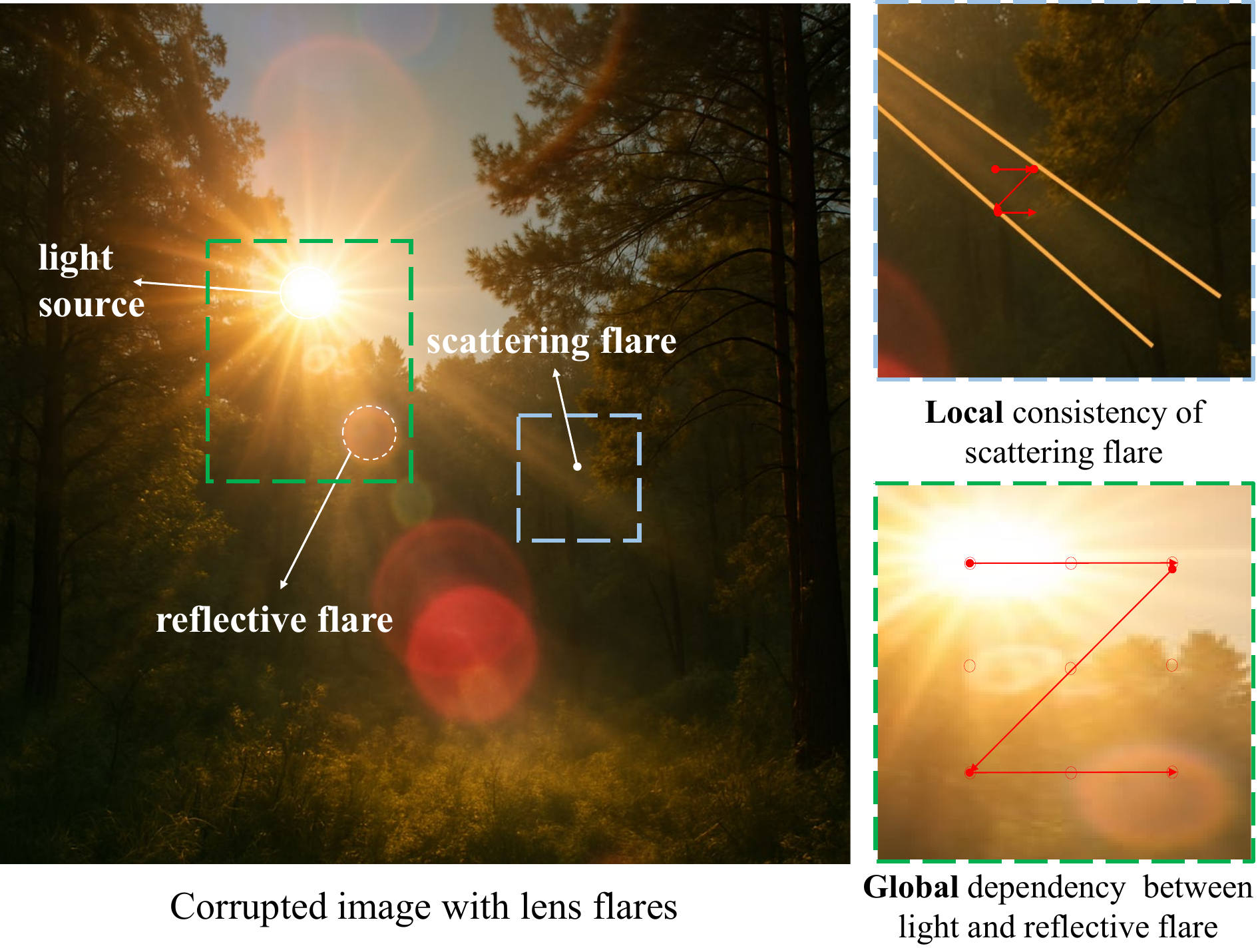}
	\caption{Global-local contextual consistency of lens flares. Exploring global-local scanning strategy is an intuitive solution to model the structure of flares and remove them.}
	\label{fig: motivation}
\end{figure}

%简单的介绍炫光任务的挑战
Lens flare manifests primarily in two forms: scattering flare and reflective flare\cite{flareunderstanding}. Scattering flare originates from the interaction between light and microscopic imperfections in the lens, resulting in diffuse streak effects. In contrast, reflective flare arises from internal reflections between lens elements, producing distinctive geometric shapes such as concentric rings or polygons. These artifacts are further amplified in low-light conditions, where the interaction between strong light sources and camera optics produces more pronounced and intricate flare patterns that weaken visual features.

%语义一致性的重要性，通过veiling flare 和 ghosting flare说明两种一致性在去眩光任务里的关键作用
In Figure~\ref{fig: motivation}, maintaining contextual consistency is crucial for flare removal tasks. 
It ensures that the restored image preserves semantic coherence of the original scene content. 
Specifically, scattering flare, characterized by localized light diffusion patterns around strong light sources, necessitates local contextual consistency to seamlessly blend the affected regions with their surroundings in terms of color saturation, texture fidelity, and edge sharpness.
In contrast, reflective flare originates from the light source and produces secondary bright spots through lens reflections. These flares maintain specific spatial relationships with their corresponding light sources. 
Thus, their removal requires coordinated processing of both the flare patterns and the associated light sources, emphasizing global semantic consistency to capture and restore the long-range relationships between light source and the induced flare patterns.
In this way, global semantic consistency facilitates detail preservation while maintaining overall scene integrity.

%现有架构CNN，Transformered不足之处——全局语义一致性难以捕捉n
However, existing deep learning architectures face significant challenges in maintaining such  
contextual consistency. CNN-based approaches~\cite{jing2021hinet,mehri2021mprnet,unet} are limited by their local receptive 
fields, making them inadequate for modeling long-range dependencies. Although Transformer-based methods~\cite{wang2022uformer, zamir2022restormer} offer enhanced global context modeling, their quadratic computational 
complexity restricts the processing of high-resolution images. While windowed self-attention mechanisms~\cite{swintransformer} attempt to mitigate this limitation, they often compromise the ability to capture the global dependencies essential for effective flare removal.

%为什么用Mamba架构
Recently, the Mamba architecture~\cite{mamba} has emerged as a promising alternative for sequence 
modeling. It employs State Space Models (SSMs) with data-dependent parameters, enabling efficient 
sequence processing with linear computational complexity. Adapting Mamba for vision tasks requires selective scan mechanisms that transform 2D image data into 1D sequences for processing. Through this mechanism, pioneering works such as ViM~\cite{vim} and VMamba~\cite{liu2024vmamba} have achieved comparable or superior performance to state-of-the-art architectures across various vision tasks. Furthermore, MambaIR~\cite{guo2024mambair} demonstrates Mamba's effectiveness in image restoration, achieving remarkable results in super-resolution, denoising, and other enhancement tasks.

%从现有scan的方式结合图片说明我们的动机。
Nevertheless, MambaIR encounters significant challenges in maintaining contextual consistency due to two primary issues:
%Flare removal differs from other restoration tasks like super-resolution or denoising, where images can be cropped into small patches (typically 64×64). In flare removal, artifacts may appear anywhere in the image, necessitating the processing of larger images (typically 512×512) to ensure that potential flare artifacts in any sub-region of the image can be effectively captured and removed.%（是这个原因吗？不是这个原因）
%这个从局部语义一致性上讲，不要讲细节，讲思路方向层面。实验啥的都别说。

1) Conventional selective scan in SSMs prioritizes broader patterns at the expense of fine-grained details, thereby overlooking the intricate relationships among local neighboring pixels. This lack of localized modeling undermines the capture of contextual dependencies critical for restoring subtle transitions and ensuring coherence within affected regions.

2) SSMs exhibit a long-term decay property, where the correlation between sequential elements diminishes as their distance increases. As illustrated in Figure~\ref{fig: motivation}, conventional selective scan strategies may map spatially adjacent flare-affected pixels to distant positions in the sequence domain. This spatial inconsistency is particularly problematic for reflective flares which often appear far from their light sources. Such spatial inconsistency prevents the effective capture of both local details and global patterns essential for high-quality flare removal.

%增加核心贡献，一句话
To address these challenges, we propose DeflareMamba, an SSM-based model specifically designed for flare removal. Our approach focuses on preserving both local spatial coherence and global structural integrity through innovative scanning strategies. Our contributions are summarized as follows:
\begin{itemize}
	\item To the best of our knowledge, this is the first work to apply Mamba for flare removal. Our experiments demonstrate that contextually consistent Mamba models can effectively address flare removal challenges, achieving promising results.
    
	\item We propose a novel hierarchical framework to simultaneously capture both local spatial coherence and global structural integrity, ensuring contextual coherence across multiple spatial scales, especially critical for handling complex flare patterns spanning various regions of the image.
    
	\item We utilize a Local-enhanced Selective Scan mechanism that effectively maintains contextual consistency between flare-affected regions and surroundings. It is specifically designed to address the inherent limitations of conventional scanning approaches in preserving spatial relationships between light sources and their associated flare artifacts.
	
	%    \item We incorporate a U-shaped network structure into our Mamba-based framework. The 
	%    downsampling operations in this structure significantly expand the receptive field of each 
	%    sequence element in SSM. We demonstrate that simply adding the U-shaped structure can 
	%    substantially improve the performance of Mamba models in flare removal tasks.%这个不是贡献
\end{itemize}
Extensive experiments on benchmark datasets demonstrate that DeflareMamba outperforms 
state-of-the-art methods in both quantitative metrics and visual quality. Furthermore, several downstream applications highlight the superiority of our method in improving both single-modal and cross-modal understanding. 

\section{Related Work}
\subsection{Flare Removal}
%\textbf{Definition} The task of nighttime flare removal aims to eliminate lens flare 
%artifacts while preserving the underlying scene content and maintaining natural image appearance. 
%These artifacts primarily manifest in two categories: scattering flares and reflective flares. 
%Scattering flares emerge from light interacting with microscopic imperfections within the lens 
%system, resulting in diffused glare patterns and distortions. Reflective flares, conversely, 
%arise from internal reflections between lens elements, producing geometric shapes such as 
%concentric rings or polygons. In nighttime scenarios, these artifacts present unique challenges 
%due to the stark contrast between bright light sources and dark backgrounds, appearing more 
%pronounced and complex than their daytime counterparts. The characteristics of these flares 
%are further influenced by lens design, aperture shape, and light source positioning, ultimately 
%leading to significant degradation of image quality and obscuration of scene details.
%
%\noindent\textbf{Existing Solutions} 
Early methods for flare removal adopted a detection-based strategy~\cite{asha2019auto,0Automated,vitoria2019automatic}, wherein flare regions are first identified based on their visual features, and subsequent restoration algorithms~\cite{1323101} are employed to remove the flares. However, these methods suffer from two primary drawbacks: they are effective only for specific flare types, and they often mistake bright objects in the image for flare artifacts.
Recent advances in data-driven approaches have considerably enhanced flare removal performance, with several innovative methods addressing the challenges associated with training data generation. Wu et al.~\cite{wu2021train} introduced a synthesis framework that leverages light source guidance for single-image flare removal. A significant breakthrough occurred with Dai et al.~\cite{dai2022flare7k}, who introduced the Flare7K dataset as the first comprehensive benchmark for nighttime flare removal research. This dataset has spurred  several follow-up studies~\cite{song2023hard,ffformer,zhou2023improving} that have further advanced flare removal techniques.Subsequently, they extended this work with Flare7K++~\cite{dai2023flare7k++}, which incorporates real-world flare data and proposes an enhanced removal pipeline, thereby achieving substantial improvements in performance.

Currently, there are only three publicly available datasets for flare removal: the dataset proposed by Wu et al.~\cite{wu2021train}, Flare7K~\cite{dai2022flare7k}, and Flare7K++~\cite{dai2023flare7k++}. While Wu's dataset only provides evaluation results on U-Net and their proposed network, the Flare7K++ dataset encompasses Flare7K and offers comprehensive benchmarking results across multiple state-of-the-art models. Therefore, we adopt Flare7K++ as our primary dataset for both training and evaluation.
\subsection{Image Restoration with Vision Mamba}
\noindent\textbf{CNN-based or Transformer-based methods} The evolution of image restoration 
techniques has witnessed a paradigm shift from traditional methods to deep learning approaches. 
CNN-based methods~\cite{dong2014learning,lepcha2023image,lim2017enhanced,zhang2018image,zamir2021multi} have demonstrated significant advantages over traditional 
approaches, offering faster training, automatic feature extraction, and superior restoration 
quality. However, these methods are inherently limited by their local receptive fields, 
constraining their ability to capture long-range dependencies.
Transformer-based architectures~\cite{liang2021swinir,wang2022uformer,zamir2022restormer} emerged as a promising solution, 
introducing self-attention mechanisms that excel at modeling global relationships in images. 
While these methods achieve impressive restoration results through their ability to capture 
long-range dependencies, they face challenges in computational efficiency, particularly when 
processing high-resolution images due to the quadratic complexity of self-attention operations. 
This limitation often necessitates the use of local-window self-attention mechanisms, resulting in an inevitable trade-off between global receptive field and computational cost.

\noindent\textbf{Vision Mamba} The Mamba architecture, based on state space models, introduces an innovative selective state update mechanism that enables efficient modeling of long-range dependencies while maintaining linear computational complexity.
% Its initial application to vision tasks was marked by VIM~\cite{vim}, which pioneered the transformation of 2D image features into 1D sequences, achieving performance comparable to Vision Transformers in image classification tasks. VMamba~\cite{liu2024vmamba} further enhanced spatial structure perception by implementing bidirectional state space models for simultaneous modeling of horizontal and vertical image features. Subsequently, MambaIR~\cite{guo2024mambair} adapted this architecture for image restoration tasks, leveraging efficient state updates for long-range feature modeling.
A critical challenge in applying the Mamba architecture to image restoration lies in the design of selective scan strategies that transform 2D images into 1D sequences amenable to processing by SSMs. While early approaches, such as VIM, utilized simple row-wise scanning to convert 2D features into 1D sequences, this method risked discarding critical spatial structural information inherent in the multi-dimensional nature of images (i.e., height, width, and channels). To overcome this limitation, several innovative scanning strategies have been proposed: VMamba introduced bidirectional scanning; LocalMamba~\cite{localmamba} developed a local scanning mechanism through image partitioning, PlainMamba~\cite{yang2024plainmamba} implemented Z-pattern scanning to preserve spatial continuity, and EfficientVMamba~\cite{pei2024efficientvmamba} proposed interval scanning with sub-image sampling to enhance computational efficiency. Despite these advancements, preserving fine details and maintaining global structural integrity in image restoration remains an active area of research.

\begin{figure*}[t]
	\centering
	\includegraphics[width=1\textwidth]{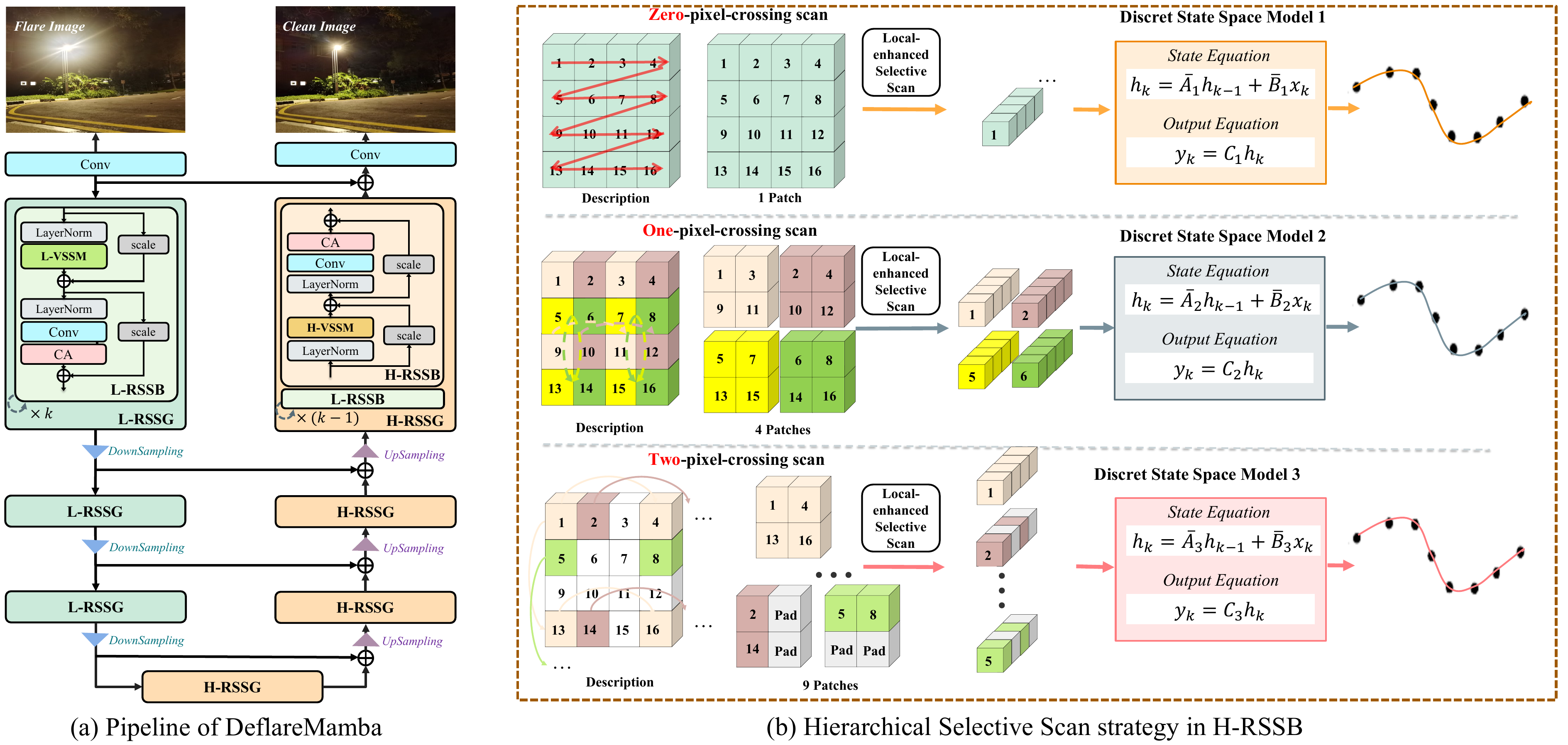}
	\caption{Overview of our DeflareMamba framework. The network adopts a U-shaped architecture 
		with Local-enhanced Residue State Space Groups in the encoding stage and Hierarchical 
		Residue State Space Groups in the decoding stage. }
	\label{fig:framework}
\end{figure*}
\section{Preliminaries}

\subsection{State Space Models}

State Space Models (SSMs)~\cite{SSM}\cite{mamba}have gained attention for their ability to model long-range dependencies with linear computational complexity relative to the input length. A continuous-time linear SSM can be described as:
\begin{equation}
\frac{d}{dt}h(t) = Ah(t) + Bx(t), \quad y(t) = Ch(t) + Dx(t),
\end{equation}
where \(h(t) \in \mathbb{R}^{N}\) is the hidden state, \(x(t)\) is the input, \(y(t)\) is the output, and \(A, B, C, D\) are learnable matrices that define the system dynamics.

For practical use in deep learning, SSMs are discretized using techniques like Zero-Order Hold (ZOH)\cite{mamba}, resulting in:
\begin{equation}
h_k = \bar{A}h_{k-1} + \bar{B}x_k, \quad y_k = Ch_k + Dx_k,
\end{equation}
with the discretized matrices given by:
\begin{equation}
\bar{A} = e^{\Delta A}, \quad \bar{B} = (\Delta A)^{-1}(e^{\Delta A} - I)B,
\end{equation}
where \(\Delta\) is the step size used in discretization.

The discrete form enables efficient parallelization using convolution:
\begin{equation}
y = x \otimes K, \quad K = (C\bar{B},\, C\bar{A}\bar{B},\, \dots,\, C\bar{A}^{L-1}\bar{B}),
\end{equation}
where \(\otimes\) denotes convolution. This formulation allows fast inference and training, making SSMs suitable for high-resolution image restoration tasks, including flare removal.

\subsection{Long-Term Decay Property of SSM}

Despite their efficiency, standard SSMs suffer from a key limitation\cite{shi2024multi}: the long-term decay property. This issue causes the influence of earlier tokens in a sequence to diminish quickly over time, leading to long-range information loss. Formally, the contribution of a past token \(m\) to a future token \(n\) (\(m < n\)) is computed as:
\begin{equation}
C_n\prod_{i=m}^{n} \bar{A}_i \bar{B}_m = C_n \bar{A}_{(m \to n)} \bar{B}_m,
\end{equation}
\text{where}
\begin{equation}
\bar{A}_{(m \to n)} = e^{\sum_{i=m}^{n} \Delta_i A}.  
\end{equation}

Since the learned \(\Delta_i A\) values are typically negative~\cite{shi2024multi}, the exponential term $e^{\sum_{i=m}^{n} \Delta_i A}$
decays rapidly as the distance \((n - m)\) increases. As a result, the effect of earlier inputs fades quickly, limiting the model’s ability to retain long-range contextual information—an important factor in image restoration tasks that involve large and spatially dispersed artifacts.

Besides, most discrete SSMs are inherently causal. Their representations are unidirectional, with information flowing only from earlier to later tokens. This design restricts later tokens from contributing to the representations of earlier ones, making it difficult to model bidirectional dependencies or global interactions effectively.
These two limitations—the long-term decay and causal structure—pose challenges for flare removal, where maintaining semantic consistency across distant and structurally connected regions is essential. Addressing these limitations is crucial for developing more capable architectures for image restoration.%具体说明 structural 的缺陷

\section{Method}

\subsection{Overall Architecture}
% As shown in Figure~\ref{fig:framework}, our DeflareMamba utilizes a U-shaped architecture, consisting groups of Local-enhanced or Hierarchical Residue State Space Groups (L-RSSB or H-RSSB). 
Our DeflareMamba framework employs a U-shaped\cite{unet} network architecture composed of several Local-enhanced and Hierarchical Residue State Space Groups (L-RSSG and H-RSSG), as illustrated in Figure~\ref{fig:framework}. In this section, we first introduce the overall U-shaped network structure and detail the composition of L/H-RSSG. Then, we elaborate on the key enhancement mechanisms in our architecture: the Local-enhanced SSM in L-RSSG, which maintains local contextual consistency; and finally, the innovative Hierarchical Selective Scan mechanism in H-RSSG, which establishes global semantic consistency essential for effective flare pattern recognition and removal.

\noindent\textbf{U-shaped architecture}.
The input resolution required for flare removal tasks is typically 512×512, which is substantially larger than the patch size processed in MambaIR\cite{guo2024mambair}. Due to the long-term decay property of SSM, the correlation between pixels weakens as their sequence distance increases. The U-shaped structure can address this limitation through two key aspects. First, downsampling operations aggregate information from surrounding pixels, enlarging the receptive field of each element in the SSM sequence. Second, reduced resolution shortens sequence lengths, enhancing correlation between distant pixels in the original image.
% Our empirical studies reveal that this constraint %%%合并一下
% typically necessitates processing significantly larger input dimensions compared to other 
% restoration tasks. Through experimentation(Section 4.2), we demonstrate that even with attempts 
% to reduce network parameters, the direct application of existing architectures continues to 
% suffer from two major drawbacks: slow convergence speed and suboptimal performance. To 
% address these fundamental challenges, we adopt a U-shaped network architecture that specifically
% targets the unique requirements of flare removal. 

Therefore, we adopt a U-shaped network architecture that specifically targets the unique requirements of flare removal. Specifically, given an input image \(I \in \mathbb{R}^{H \times W \times 3}\), we first perform coarse feature extraction through a convolutional layer to project the image into a higher-dimensional embedding 
space \(\mathbb{R}^{H \times W \times C}\). 
%加了一句衔接的话
The extracted feature maps are subsequently processed by a U-shaped network architecture that comprises an encoding stage and a decoding stage.

1) In the encoding stage, we employ Local-enhanced Residue State Space Groups 
(L-RSSG) for hierarchical feature extraction. Within the \(l\)-th L-RSSG, features first pass through 
\(l\) consecutive Local-enhanced Residue State Space Blocks (L-RSSB) for deep feature extraction. 
Each L-RSSB is designed to capture fine-grained local details. An additional convolutional layer is introduced 
at the end of each group to refine the extracted features, followed by an element-wise summation 
with the input through a residual connection to facilitate gradient flow and prevent feature 
degradation during deep propagation. 

2) In the decoding stage, we adopt Hierarchical Residue State Space 
Groups (H-RSSG) to process features at different scales. The structure of H-RSSG closely mirrors 
that of L-RSSG, beginning with \(l-1\) L-RSSB blocks and concluding with a convolutional layer 
and residual connection. The key distinction lies in the final block, where we replace the 
standard L-RSSB with our Hierarchical RSSB (H-RSSB) to enable comprehensive hierarchical feature 
extraction. This novel H-RSSB incorporates varied stride sampling patterns to establish long-range 
pixel correlations, effectively capturing both local details and global structural information 
necessary for high-quality flare removal.

%%%待定，等Wu的实验跑完
Following the pipeline established in Flare7k++\cite{dai2023flare7k++}, the output features from our U-shaped network 
are projected through a convolutional layer to produce a 6-channel output tensor 
\(O \in \mathbb{R}^{H \times W \times 6}\). The first three channels \(O_{1:3}\) represent the 
flare-free image, serving as the primary output for the flare removal task, while the remaining 
three channels \(O_{4:6}\) contain the predicted flare image, which is utilized in computing 
auxiliary losses to facilitate better network optimization.

%注意全文的一致
\noindent\textbf{Local-enhanced/Hierarchical Residue State Space Blocks}. The basic blocks L-RSSB and H-RSSB in our architecture follow the structure of RSSB proposed in MambaIR~\cite{0Automated}, but incorporate our improved 
selective scan mechanism to enhance the model's capability in capturing global-local contexture consistency. 
Similar to the Transformer architecture's flow of Norm → Attention → Norm → MLP, our H/L-RSSB 
replaces the self-attention mechanism with our proposed H/L-VSSM (Hierarchical/Local-enhanced 
Vision State Space Module).

Given an input feature map \(X \in \mathbb{R}^{H \times W \times C}\), the H/L-VSSM first applies layer normalization:

\begin{equation}
	X_1 = \textit{LN}(X),
\end{equation}

As shown in Figure~\ref{fig:framework}, the normalized features are then processed through two parallel branches. 
The main branch incorporates a depth-wise convolution (DWConv) for local feature extraction, followed by our 
improved two-dimensional state space model \(\Psi_{SSM}\):

\begin{equation}
	X_m = \Phi_1(\Psi_{SSM}(\sigma(\textit{DWConv}(X_1)))),
\end{equation}

where \(\Psi_{SSM}\) represents either the Local-Enhanced SSM (L-SSM) in the encoder path or Hierarchical SSM 
(M-SSM) in the decoder path, as illustrated in the detailed block diagrams. The skip branch applies a simple 
linear projection after SiLU activation:

\begin{equation}
	X_s = \sigma(\Phi_2(X_1)),
\end{equation}

Following the design pattern shown in the block diagrams, the outputs from both branches are combined and 
processed through a final layer normalization and projection:

\begin{equation}
	Y = \Phi_3(\textit{LN}(X_m + X_s)),
\end{equation}

This dual-branch architecture is implemented in both Local Residual State Space Blocks (L-RSSB) and Hierarchical 
Residual State Space Blocks (H-RSSB), with the key difference lying in their selective scan mechanisms in the 
main branch. The L-VSSM employs Local-enhanced Selective Scan for preserving local spatial relationships, while 
H-VSSM incorporates Hierarchical Selective Scan to enable hierarchical feature processing through varied sampling patterns. 

\begin{algorithm}[h]
	\caption{Local-Enhanced SS2D}
	\label{alg:local-enhanced-ss2d}
	\begin{algorithmic}[1]
		\STATE \textbf{Inputs:} Feature map $X \in \mathbb{R}^{H \times W \times C}$
		\STATE \textbf{Hyperparameters:} Window size $(w_h, w_w)$
		\STATE Initialize output: $Y = \mathbf{0} \in \mathbb{R}^{H \times W \times C}$
		\STATE Create directional variants: $X_{dir} = [X, X^T, X^F, (X^T)^F]$
		\FOR{$i = 0$ to $3$}
		\STATE Partition into windows: $X_{win}^i = \text{Partition}(X_{dir}^i, w_h, w_w)$
		\STATE Convert to sequence: $S^i = \text{Flatten}(X_{win}^i)$
		\STATE Process with SSM: $\hat{S}^i = \text{SSM}_i(S^i)$
		\STATE Restore window structure: $Y_{win}^i = \text{Reshape}(\hat{S}^i, w_h, w_w)$
		\STATE Apply inverse transform: $Y^i = \text{Inverse}_i(Y_{win}^i)$
		\STATE Accumulate results: $Y = Y + Y^i$
		\ENDFOR
		\STATE \textbf{Return:} $Y$ 
	\end{algorithmic}
\end{algorithm}

\subsection{Local-Enhanced SSM}
%%暂时删掉，要有更好的衔接
% While effective flare removal benefits from the global context captured by our hierarchical architecture, 
% it equally requires precise attention to localized flare details. Flare artifacts typically manifest as 
% concentrated luminance anomalies in specific image regions, exhibiting distinctive spatial patterns and 
% intensity distributions with varying degrees of transparency, color shifts, and boundary diffusion 
% characteristics. These localized phenomena often present complex optical interactions that require fine-grained 
% processing to accurately identify and neutralize without compromising the underlying image content. 
% To address these requirements, we implement Local-Enhanced SS2D, which augments the standard 
% 2D Selective Scan (SS2D) mechanism with localized processing capabilities. 

% The hierarchical features extracted by our hierarchical architecture serve as direct inputs to our 
% Local-Enhanced SS2D blocks. The Local-Enhanced SS2D methodology begins by strategically partitioning 
% these scale-specific feature maps into a grid of non-overlapping windows before sequence linearization. 
% We then apply 2D-Selective-Scan operations across these windows to fully preserve spatial structural 
% features. This approach maintains local spatial relationships while enabling global context integration.
In the context of flare removal, restoring damaged regions requires effective utilization of local semantic consistency. However, in conventional selective scan sequences, spatially adjacent pixels in the original image may be positioned far apart in the sequence. Due to the inherent distance-decay property of State Space Model\cite{shi2024multi}, this can lead to significant attenuation of local contextual information. The local-scan approach proposed by \cite{localmamba}, which processes and concatenates image patches separately, effectively addresses this limitation. Building upon this foundation, we further enhance the mechanism by introducing multi-directional scanning patterns to maximally preserve the spatial structural information of the original image.

Specifically, given a scale-specific feature map $X \in \mathbb{R}^{H\times W\times C}$, we divide it into 
$M\times N$ non-overlapping windows of size $win_h\times win_w$, where $M=H/win_h$ and $N=W/win_w$. Within each
window, pixels are scanned in raster order from left to right and top to bottom, generating local subsequences 
that maintain neighborhood relationships. These subsequences are then concatenated according to their window 
positions, forming a complete sequence that preserves both local structure and global arrangement.

To maximize the preservation of inherent spatial structures in the original image, we adopt the approach \cite{liu2024vmamba} 
which processes the feature map from multiple perspectives. Specifically, we can transpose the aforementioned 
hierarchical sub-images and further apply reverse or non-reverse operations to the transposed or non-transposed 
local image sequences, thereby obtaining scan sequences from four different directions. These four directional 
sequences are processed by four separate SSM models with different parameters. Finally, the processed sequences 
are restored to the original image shape, and these four feature maps are simply added together. The detailed 
algorithm is presented in Algorithm~\ref{alg:local-enhanced-ss2d}.

This approach ensures that local semantic consistency is effectively preserved during the flare removal process, leading to more accurate restoration of flare-affected regions. Detailed ablation study results are presented in Section 4.1.

\subsection{Hierarchical Selective Scan Mechanism}
%introduction借鉴
Hierarchical mechanism plays a crucial role in image restoration~\cite{hierarchical}.  Effective flare removal requires both the utilization of local textural details and the comprehension of flare patterns that may manifest across broader spatial contexts. To effectively capture and process these hierarchical features, we propose Hierarchical Selective Scan mechanism at decoding stage.
% Specifically, this module implements a specialized hierarchical mechanism designed for State Space Models (SSMs), which circumvents the need for additional downsampling operations that could potentially result in information degradation.

The foundation of our approach lies in a key property of discrete SSMs: the correlation between two elements in a sequence decreases as their distance increases~\cite{shi2024multi}. Inspired by this property, our Hierarchical Selective Scan mechanism implements different stride sampling patterns during the selective scan process, generating multiple sub-sequences at each scale. In these sub-sequences, pixels that were originally distant in the input image become closer, thereby establishing stronger connections through SSM processing. After the SSM operation, these sub-sequences are reshaped back to the original feature map dimensions, yielding scale-specific feature representations that capture long-range dependencies without resolution loss.

% For a scale level $k$, we perform pixel sampling with a stride of $k$ pixels in both horizontal and vertical directions, resulting in a sub-sequence. Since there are $k^2$ possible starting positions in the original image, we obtain $k^2$ sub-sequences for each scale level $k$. These sub-sequences are processed by the scale-specific Local-Enhanced SS2D model (detailed in Section 3.3), where the model parameters are shared among all sub-sequences within the same scale level. After processing, the elements in each sub-sequence are mapped back to their corresponding positions in the original feature map, yielding the scale-specific feature representation. 

To formalize this Hierarchical Selective Scan mechanism, we define $i$-th level sub-image $\mathcal{S}_{ik}$ of feature map $\mathcal{F} \in \mathbb{R}^{H \times W \times C}$ as:

\begin{equation}
	(\mathcal{S}_{ik})_{h,w} = 
	\begin{cases}
		\mathcal{F}_{h^{'},w^{'}}, & h^{'} \leq H-1, w^{'} \leq W-1 \\[8pt]
		0, & \text{else}
	\end{cases}
\end{equation}

\noindent where the mapped coordinates $(h',w')$ are computed as:
\begin{equation}
    \begin{gathered}
        h^{'} = h_k + 2^i \cdot h,\quad w^{'} = w_k + 2^i \cdot w, \\
    \end{gathered}
\end{equation}

\noindent with $h_k,w_k\in \{0,1,...,\lceil H/2^i \rceil-1\}$ being the top-left coordinate of the $k$-th sub-image, where $k\in \{0,1,...,2^i-1\} $ is the sub-image index and $h,w$ are the spatial coordinates in the sub-image. To facilitate efficient batch processing in our SSM operations, all sub-images are zero-padded to match the dimensions of the largest sub-image.

Each sub-image is then processed by our Local-enhanced SSM module to obtain processed sub-features:

\begin{equation}
	\hat{\mathcal{S}}_{ik} = LSSM_i(\mathcal{S}_{ik}),
\end{equation}

\noindent where $LSSM_i$ denotes the Local-enhanced SSM operation at scale level $i$. These processed sub-features are then reverse-mapped to reconstruct scale-specific features:

\begin{equation}
	\widetilde{\mathcal{F}}_i = \text{Reverse}\left(\left\{\hat{\mathcal{S}}_{ik}\right\}_{k=0}^{2^i-1}\right),
\end{equation}

\noindent where $\text{Reverse}$ is the function that maps processed sub-images back to the original feature space. Finally, we integrate features across all $K$ scale levels to obtain the enhanced feature representation:

\begin{equation}
	\widetilde{\mathcal{F}} = \frac{1}{K}\sum_{i=1}^{K}\widetilde{\mathcal{F}}_i,
\end{equation}

Notably, our approach operates in a divide-and-conquer-like manner on the original feature map, fundamentally differing from previous methods that rely on resolution reduction.

\begin{figure}
    \centering
    \includegraphics[width=1.0\linewidth]{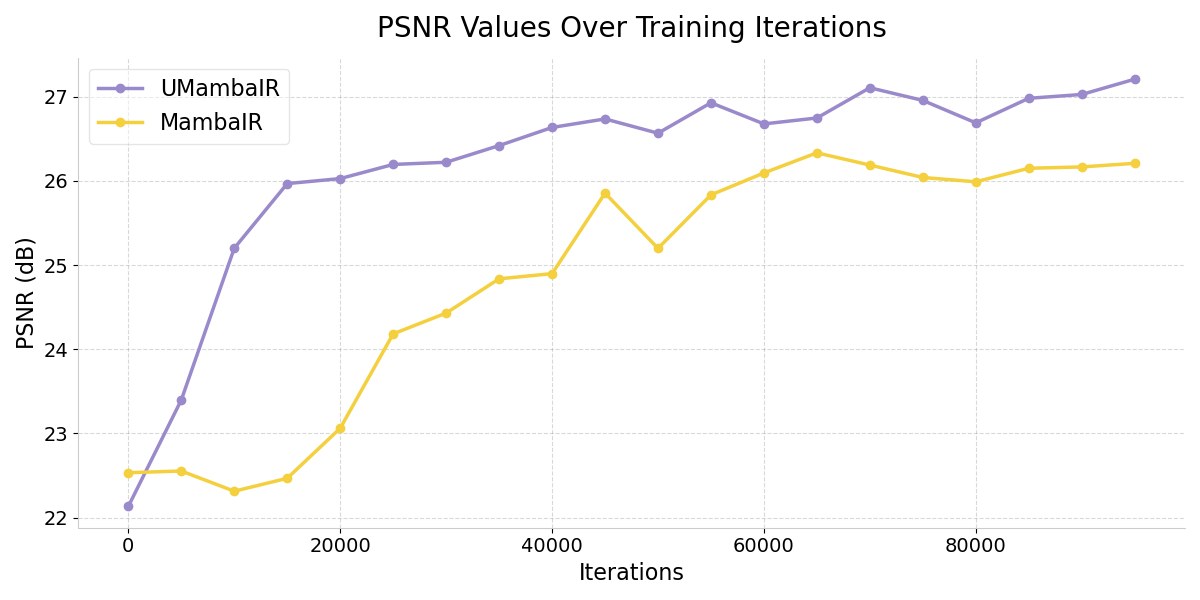}
    \caption{PSNR values over training iterations for U-shaped MambaIR and MambaIR architectures. The U-shaped structure demonstrates faster convergence and better performance compared to the MambaIR.}
    \label{fig:convergence_comparison}
\end{figure}

\begin{figure*}[t]
	\centering
	\includegraphics[width=\textwidth,height=0.47\textheight]{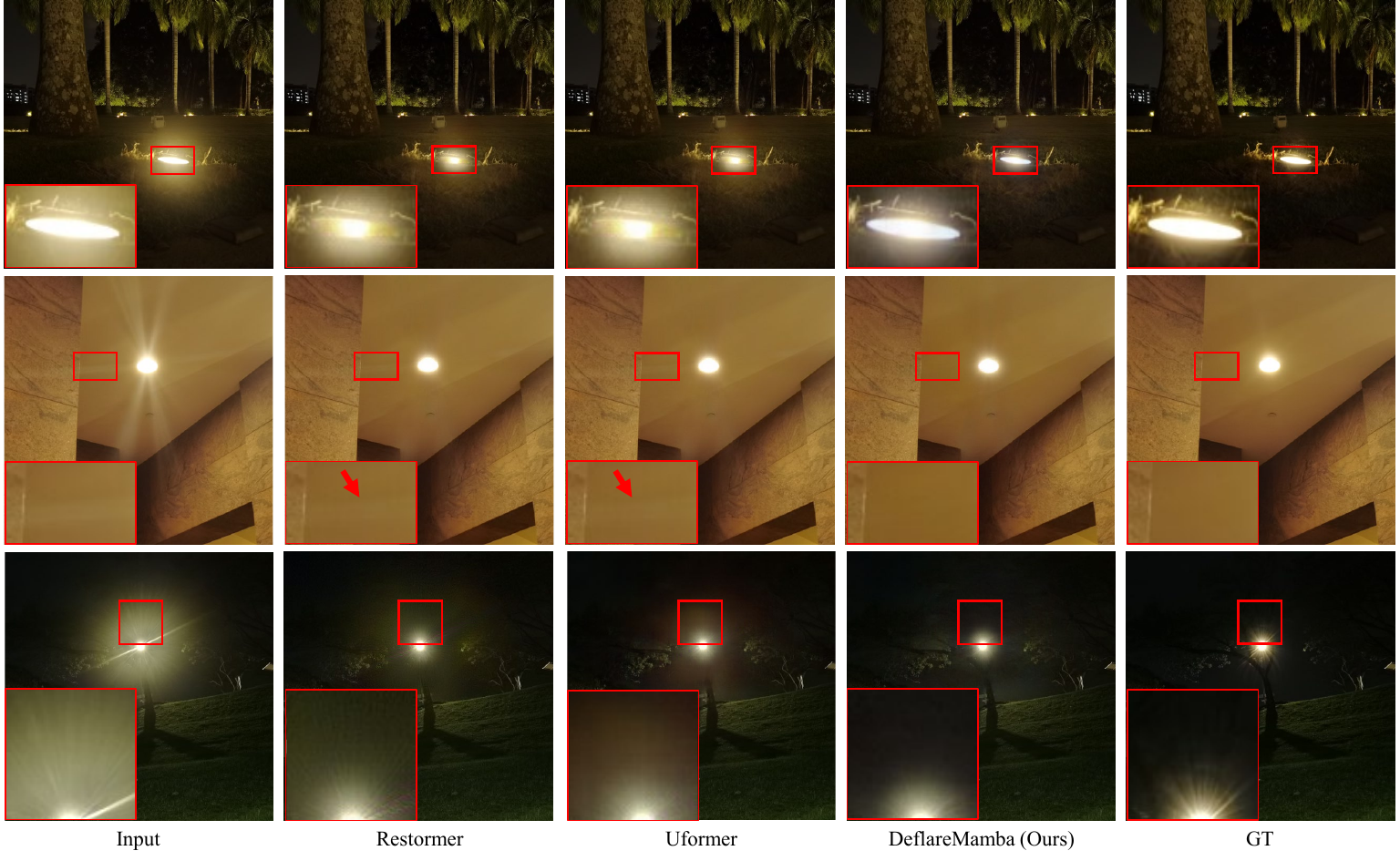}
	\caption{Qualitative comparison of flare removal results. From left to right: Input image with lens flares, Restormer, Uformer, DeflareMamba (Ours), and Ground Truth (GT). Our method better preserves image details near light sources and achieves cleaner removal of streak artifacts, leading to more visually pleasing results that closely match the ground truth.}
	\Description{Six images arranged horizontally, demonstrating the comparison of different methods for nighttime lens flare removal.}
	\label{fig:visual_comparison}
\end{figure*}

\begin{table}[t]
	\caption{Ablation Study on Local-Enhanced and Hierarchical Components}
	\label{tab:ablation}
	\begin{tabular}{lcc}
		\toprule
		Model Configuration & PSNR & SSIM \\
		\midrule
		Baseline & 27.354 & 0.894 \\
		+ Local-enhanced SS2D & 27.627 & 0.896 \\
		+ Hierarchical Selective Scan & 27.778 & 0.899 \\
		\bottomrule
	\end{tabular}
\end{table}

%%%增加了两列，同时将psnr保留了两位小数
\begin{table*}[t]
\caption{Quantitative Comparison with State-of-the-art Methods}
\label{tab:comparison}
\resizebox{1.0\textwidth}{!}{
\begin{tabular}{c|ccccccccc}
\toprule
\multirow{2}{*}{Metric} & \multicolumn{9}{c}{Network trained on Flare7K++} \\
\cmidrule{2-10}
& Input & U-Net\cite{wang2022uformer} & HINet\cite{jing2021hinet} & MPRNet*\cite{mehri2021mprnet} & Restormer*\cite{zamir2022restormer} & Uformer\cite{wang2022uformer} & Difflare \cite{difflare} & Kopt.el\cite{kopt} & 
DeflareMamba (Ours) \\
\midrule
PSNR↑    & 22.56 & 27.19 & 27.55 & 27.04 & 27.60 & 27.63 & 26.06 & 27.66 & \textbf{27.78} \\
SSIM↑    & 0.857  & 0.894  & 0.892  & 0.893  & 0.897  & 0.894 & 0.898 & 0.897 & \textbf{0.899} \\
\bottomrule
\end{tabular}
}
\end{table*}

\section{Experiments}
\subsection{Implementation Details}
We train our network on the Flare7K++ dataset\cite{dai2023flare7k++}. For flare synthetic process, we follow the strategy 
as Flare7K++, where paired flare-corrupted and flare-free images are generated on-the-fly by 
sampling background images from the 24K Flickr dataset\cite{flickr24k}. 
% The flare images and corresponding light sources are sampled from both Flare7K\cite{dai2022flare7k} and Flare-R datasets with equal probability. 
To enhance the diversity and robustness of the training data, we apply various augmentation techniques 
including random rotation, translation, shearing, scaling, blurring, and color adjustments. 
All these data augmentation settings are kept identical to those in Flare7K++ to ensure fair 
comparison.

Our network training framework and settings are the same as Flare7k++.
Specifically, input images are center-cropped to 512×512, and the model is trained with 
a batch size of 2 using the Adam \cite{kingma2014adam} optimizer with a learning rate of 1e-4 for 300k iterations. 
The backbone outputs six channels, where the first three channels represent the flare-free 
image and the latter three channels correspond to the flare image. These outputs are used 
to compute the loss function:

\begin{equation}
	\hat{I_0}, \hat{F} = \mathcal{N}(I),
\end{equation}
\begin{equation}
	L(\hat{y}, y) = L_1(\hat{y}, y) + L_{vgg}(\hat{y}, y),
\end{equation}
\begin{equation}
	L_{rec} = |I - Clip(\hat{I_0} \oplus \hat{F})|,
\end{equation}
\begin{equation}
	L_{total} = w_1L(\hat{I_0}, I_0) + w_2L(\hat{F}, F) + w_3L_{rec},
\end{equation}
where $\mathcal{N}$ takes the flare-corrupted image $I$ as input and outputs the predicted flare-free image $\hat{I_0}$ and flare component $\hat{F}$. Both outputs are compared with their ground truth counterparts ($I_0$ and $F$) using loss function $L$ that combines L1 and VGG \cite{vgg} perceptual losses, while $L_{rec}$ measures the reconstruction error between the input image and the recomposed image from predictions. The total loss $L_{total}$ is a weighted combination. We empirically set $w_1 = w_2 = w_3 = 1$, which is consistent with the settings in Flare7K++.

\subsection{Ablation Study}

\noindent\textbf{Effect of U-shaped Network Architecture.} Our framework adopts a U-shaped architecture to gradually expand the receptive field of SSM elements while reducing sequence lengths. For ablation studies on U-shaped architecture, we use MambaIR\cite{guo2024mambair} as the baseline. We then introduce U-shaped MambaIR by incorporating downsampling and upsampling operations into MambaIR. Experiments demonstrate that this design effectively maintains contextual consistency throughout the network. 

As shown in Figure~\ref{fig:convergence_comparison}, U-shaped MambaIR achieves faster convergence compared to the baseline MambaIR\cite{guo2024mambair}. During early training (first 20k iterations), the U-shaped network gains nearly 3dB advantage. After complete training, it maintains approximately 1dB higher PSNR, confirming that the U-shaped structure effectively addresses SSM limitations.

\noindent\textbf{Effect of Local-Enhanced and Hierarchical Mechanism.}
As discussed earlier, flare removal tasks require both global perception and local focus 
for effective restoration. We select the aforementioned U-shaped MambaIR as our baseline.
 We first replace the standard SS2D in baseline with 
Local-enhanced SS2D. Then, we further introduce the Hierarchical Selective Scan mechanism to the model. 
Table~\ref{tab:ablation} presents the quantitative results.

The ablation study demonstrates significant PSNR improvements from our proposed components. 
The Local-enhanced SS2D improves the baseline PSNR by 0.273dB, indicating its effectiveness 
in preserving local spatial relationships and fine-grained details. Furthermore, the Hierarchical 
Selective Scan mechanism adds another 0.151dB gain, confirming its ability to establish 
stronger long-range dependencies across different scales. Together, these components achieve 
a total improvement of 0.424dB while maintaining consistent gains in SSIM\cite{ssim}.

\subsection{Comparison with SOTAs}
To evaluate the effectiveness of our proposed DeflareMamba, we conduct comprehensive 
comparisons with state-of-the-art networks trained on Flare7K++. For a fair comparison, 
we evaluate all methods on the same test set and report standard evaluation metrics 
including PSNR and SSIM\cite{ssim}. Methods marked with asterisks (*) indicate reduced parameter 
configurations, with implementation details following Flare7K++ ~\cite{dai2023flare7k++}.

\begin{figure}[!t]
	\centering
	\includegraphics[width=1\linewidth]{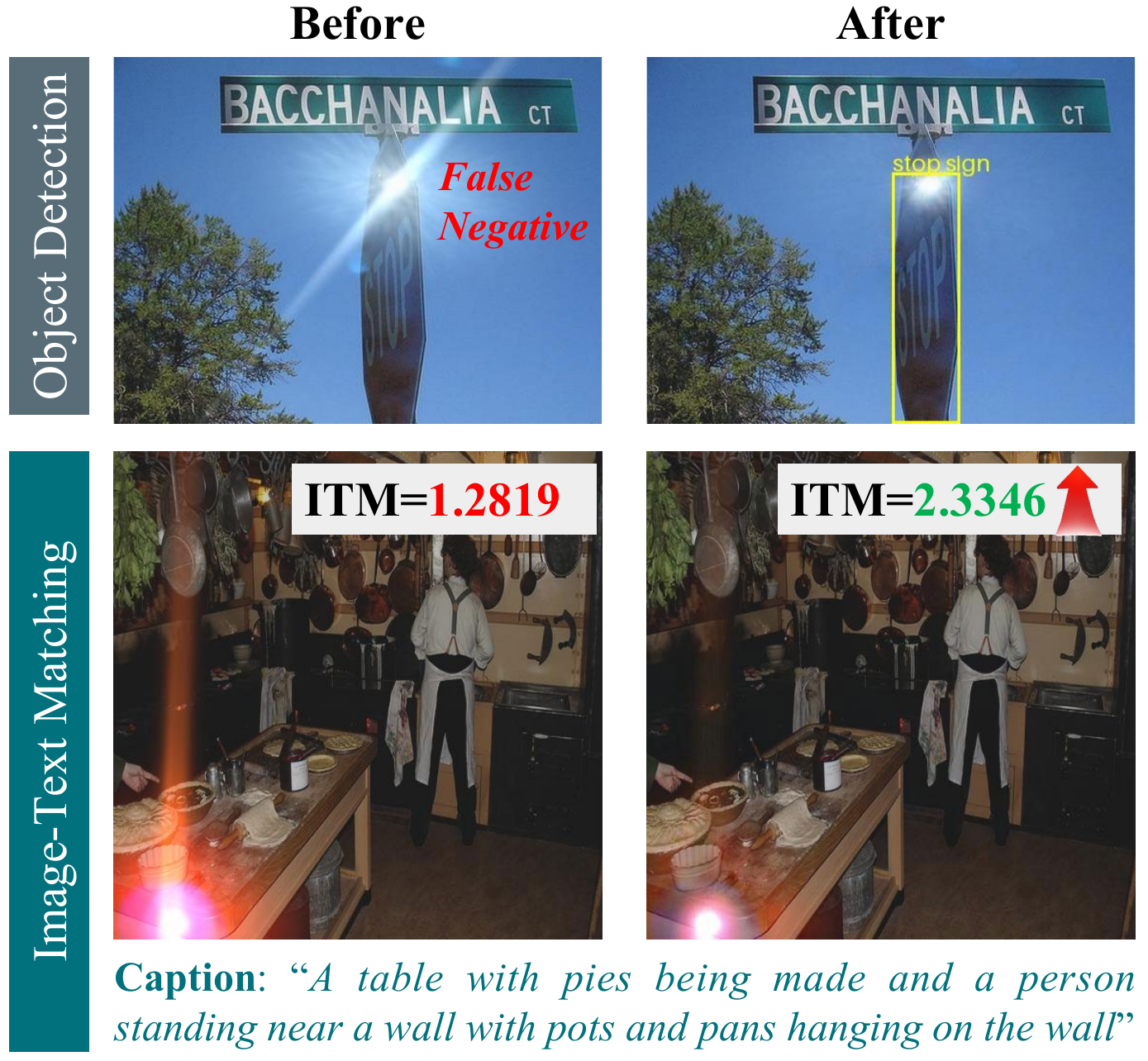}
	\caption{Visual results of object detection and image-text matching before/after flare removal. Our method successfully prevents the false negative detection of the street sign and improves the image-text matching score.}
	\label{fig:downstream}
\end{figure}

As shown in Table~\ref{tab:comparison}, DeflareMamba achieves the best performance among all compared methods. Specifically, it outperforms the previous best performer Uformer\cite{wang2022uformer} by 0.145dB in PSNR. For SSIM~\cite{ssim}, our method achieves 0.899, which is 0.005 higher than Uformer (0.894) and 0.002 better than the previous best performer Restormer \cite{zamir2022restormer}(0.897). These improvements demonstrate the effectiveness of our approach in preserving both structural and perceptual image quality. Visual comparisons in Figure~\ref{fig:visual_comparison} further reveal that our method restores more distorted details and reduces flare artifacts more completely. More implementation details and comparison results are available in supplementary materials.

\begin{table}[!t]
\small  % 或者用 \small，看哪个大小更合适
\caption{Object Detection Performance (mAP)}
\label{tab:detection}
\resizebox{1.0\linewidth}{!}{
\begin{tabular}{l|ccc}
\toprule
Data & Faster R-CNN~\cite{fasterrcnn} & Deform. DETR~\cite{ddetr} & YOLO8x~\cite{yolov8} \\
\midrule
w/ flare & 28.15 & 36.20 & 38.15 \\
Uformer & 29.13 & 37.35 & 39.04 \\
\textbf{DeflareMamba} & \textbf{29.35} &  \textbf{37.59} & \textbf{39.36} \\
\bottomrule
\end{tabular}
}
\end{table}

\begin{table}[!t]
\small
\caption{Vision-Language Alignment Performance}
\label{tab:matching}
\begin{tabular}{l|c|cccc}
\toprule
\multirow{2}{*}{Data} & CLIP\cite{clip} & \multicolumn{4}{c}{BLIP\cite{blip}} \\
\cmidrule{2-6}
& CS & CS & ITM & TR@1 & TR@5 \\
\midrule
w/ flare & 28.57 & 45.31 & 117.56 & 72.2 & 90.1 \\
Uformer & 29.14 & 46.37 & 127.79 & 73.1 & 91.0 \\
\textbf{DeflareMamba} & \textbf{29.22} & \textbf{46.50} & \textbf{128.45} & \textbf{73.4} & \textbf{91.5} \\
\bottomrule
\end{tabular}
\end{table}

\subsection{Performance on Downstream Tasks}
To evaluate our method's impact on downstream tasks, we synthesize test images by applying flare effects from Flare7K++ to the COCO \cite{coco} validation set. We then compare the performance of pre-trained models on these flare-corrupted images before and after processing with Uformer \cite{wang2022uformer} and our DeflareMamba.

We evaluate our method using pre-trained models for object detection and vision-language alignment tasks on the COCO dataset. For object detection, we measure the mean Average Precision (mAP). For vision-language alignment, we assess both the semantic alignment quality through similarity scores and text-to-image retrieval performance using Top-k recall rates (TR@k). As shown in Table~\ref{tab:detection} and Table~\ref{tab:matching}, our method consistently outperforms both flare-corrupted inputs and Uformer across all metrics, demonstrating its effectiveness in preserving semantic information for downstream tasks. More details are provided in the supplementary materials.
% 再提及一下可视化结果中，各自的下游任务方法是什么。

\section{Conclusion and Discussion}
% In this paper, we present DeflareMamba, a novel nighttime flare removal framework based on hierarchical state space models. In existing data-driven flare removal architectures, CNN-based approaches suffer from restricted receptive fields, while Transformer-based methods with windowed self-attention constrain global context modeling, requiring a careful balance between global perception and computational complexity. To address this problem, we pioneer the application of Mamba architecture to the flare removal domain and propose task-specific adaptations. To enhance the receptive field of each token in the SSM sequence while reducing complexity, we implement a U-shaped network structure. Furthermore, to strengthen the model's hierarchical capability, we propose a novel hierarchical mechanism specifically designed for SSM models, which achieves hierarchical processing by sampling sub-images at different scales with varying pixel intervals, eliminating the need for downsampling as required in previous methods. Meanwhile, to enhance local perception and better process flare-affected regions, we introduce the Local-Enhanced SS2D mechanism that effectively captures local spatial information. We train our model on the Flare7K++ dataset, and experimental results demonstrate that DeflareMamba outperforms state-of-the-art methods in both quantitative metrics and visual quality, particularly excelling in challenging scenarios with complex flare patterns and varying illumination conditions, making it a promising solution for nighttime flare removal.
In this paper, we propose DeflareMamba, the first Mamba-based architecture for flare removal that effectively maintains global-local contextual consistency. We adopt a U-shaped architecture to mitigate the long-term decay issue in SSMs by reducing sequence lengths while aggregating neighboring pixel information. To better preserve local semantic consistency, we employ the Local-enhanced SS2D mechanism with multi-directional scanning patterns. Moreover, we propose a novel Hierarchical Selective Scan mechanism that captures hierarchical contextual information through cross-level pixel sampling. Extensive experiments demonstrate that our method not only outperforms state-of-the-art flare removal approaches but also enhances both object detection performance and vision-language semantic understanding by reducing flare-induced information distortion.

\noindent\textbf{Discussion.} While our method demonstrates significant improvements in flare removal, there are several directions for future research. First, the complete removal of long streak artifacts in scattering flares remains challenging, suggesting the need for improved architectures or data augmentation strategies. Second, our current single-block hierarchical feature processing could be distributed across multiple blocks while maintaining the total block count, potentially reducing computational complexity.

\section*{Acknowledgments}
This work was supported in part by the National Natural Science Foundation of China under Grant 62202056, and the Fundamental Research Funds for the Central Universities under Grant 2243100002.
%%
%% The next two lines define the bibliography style to be used, and
%% the bibliography file.
\bibliographystyle{ACM-Reference-Format}
\bibliography{sample-base}

\end{document}